%% file: main_full.tex
\documentclass[conference]{IEEEtran}
\usepackage{times}

\usepackage[numbers]{natbib}
\usepackage{multicol}

\IEEEoverridecommandlockouts                              % This command is only needed if 
\usepackage{graphics} % for pdf, bitmapped graphics files
\usepackage{epsfig} % for postscript graphics files
\usepackage{mathptmx} % assumes new font selection scheme installed
\usepackage{times} % assumes new font selection scheme installed
\DeclareMathAlphabet{\mathcal}{OMS}{cmsy}{m}{n}
\usepackage{subfig}
\usepackage{algorithm}
\usepackage{algpseudocode}
\usepackage{amsmath} % assumes amsmath package installed
\usepackage{amssymb}  % assumes amsmath package installed
\usepackage{hyperref}
\hypersetup{
colorlinks=true,
linkcolor=black,
hyperindex=true,
citecolor=red}
\newcommand{\algname}{\mbox{IJP}}
\newcommand{\fullalgname}{\mbox{Interactive Joint Planning}}

\usepackage{color}
\usepackage{pifont}
\usepackage{multirow}

\DeclareMathOperator{\bue}{\mathbf{u}_e}
\DeclareMathOperator{\bxe}{\mathbf{x}_e}
\DeclareMathOperator{\buo}{\mathbf{u}_o}
\DeclareMathOperator{\bxo}{\mathbf{x}_o}
\DeclareMathOperator{\xref}{\mathbf{x}_{\text{ref}}}
\DeclareMathOperator{\xpred}{\mathbf{x}_{\text{pred}}}
\DeclareMathOperator{\jref}{\mathcal{J}_{\text{ref}}}
\DeclareMathOperator{\jdev}{\mathcal{J}_{\text{dev}}}
\DeclareMathOperator{\ju}{\mathcal{J}_{\text{u}}}

\usepackage{algorithmicx}
\usepackage{algorithm}
\usepackage{algpseudocode}
\usepackage{xcolor}
\algdef{SE}[DOWHILE]{Do}{DoWhile}{\algorithmicdo}[1]{\algorithmicwhile\ #1}%

\newtheorem{definition}{Definition}
\newtheorem{remark}{Remark}

\newtheorem{thm}{Theorem}
\newtheorem{lemma}{Lemma}

\usepackage{xcolor}
\newcommand{\comment}[1]{{\color{blue}Comment: #1}} 
 
\renewcommand{\comment}[1]{} % disable
\title{\LARGE \bf
Interactive Joint Planning for Autonomous Vehicles
}

\author{Yuxiao Chen$^{1}$, Sushant Veer$^{1}$, Peter Karkus$^{1}$, and Marco Pavone$^{2}$
\thanks{$^{1}$The authors are with NVIDIA Research, NVIDIA Corporation, 2788 San Tomas Expressway, Santa Clara, CA.
        {\tt\small \{yuxiaoc,sveer,pkarkus\}@nvidia.com}}
\thanks{$^{2}$Marco Pavone is with the Department of Aeronautics and Astronautics, Stanford University, and with NVIDIA Research.
        {\tt\small \{pavone@stanford.edu, mpavone@nvidia.com\}}}%
}

\begin{document}

\maketitle
\thispagestyle{empty}
\pagestyle{empty}

%%%%%%%%%%%%%%%%%%%%%%%%%%%%%%%%%%%%%%%%%%%%%%%%%%%%%%%%%%%%%%%%%%%%%%%%%%%%%%%%
\begin{abstract}
In highly interactive driving scenarios, the actions of one agent greatly influences those of its neighbors. Planning safe motions for autonomous vehicles in such interactive environments, therefore, requires reasoning about the impact of the ego's intended motion plan on nearby agents' behavior. Deep-learning-based models have recently achieved great success in trajectory prediction and many models in the literature allow for ego-conditioned prediction. However, leveraging ego-conditioned prediction remains challenging in downstream planning due to the complex nature of neural networks, limiting the planner structure to simple ones, e.g., sampling-based planner.
Despite their ability to generate fine-grained high-quality motion plans, it is difficult for gradient-based planning algorithms, such as model predictive control (MPC), to leverage ego-conditioned prediction due to their iterative nature and need for gradient. We present \fullalgname (\algname{}) that bridges MPC with learned prediction models in a computationally scalable manner to provide us the best of both the worlds. 
% model predictive controller that joint optimizes over the future motion of the ego vehicle together with the adjacent agents. 
In particular, \algname{} jointly optimizes over the behavior of the ego and the surrounding agents and leverages deep-learned prediction models as prediction priors that the join trajectory optimization tries to stay close to.
Furthermore, by leveraging homotopy classes, our joint optimizer searches over diverse motion plans to avoid getting stuck at local minima. Closed-loop simulation result shows that \algname{} significantly outperforms the baselines that are either without joint optimization or running sampling-based planning.
\end{abstract}

\section{Introduction}\label{sec:intro}
\input{introduction.tex}

\section{Related Works}\label{sec:related-work}
\input{related_works.tex}
\section{Free-end Homotopy Classes}\label{sec:homotopy}
\input{homotopy_full.tex}
\section{Method}\label{sec:method}

\input{method_full.tex}
\section{Simulation setup and results}\label{sec:result}
\input{result_full.tex}

\section{Conclusion and discussion}\label{sec:conclusion}
\input{conclusion.tex}
%%%%%%%%%%%%%%%%%%%%%%%%%%%%%%%%%%%%%%%%%%%%%%%%%%%%%%%%%%%%%%%%%%%%%%%%%%%%%%%%

% \section{Conclusion}

% \section*{Acknowledgement}

\appendix
\input{appendix}

\bibliographystyle{IEEEtran}
\bibliography{bibliography}

\renewcommand{\baselinestretch}{0.97}

\end{document}

%% file: introduction.tex
A cornerstone for safe motion planning for autonomous vehicles is the ability to reason about interactions between the ego vehicle and other traffic agents, such as human-driven vehicles and pedestrians.
% Motion planning for autonomous vehicles requires reasoning about interaction between the ego vehicle and the adjacent agents such as human-driving vehicles and pedestrians. 
A standard approach to deal with interactive scenarios is to leverage prediction models---heuristic \cite{liebnerBaumannEtAl2012} or data-driven \cite{BansalKrizhevskyEtAl2019,CuiSergioEtAl2021}---to generate predictions of the traffic agents' future motions and plan the ego motion accordingly. In particular, various deep-learned prediction models now represent the state of the art in prediction \cite{SalzmannIvanovicEtAl2020,ChenIvanovicEtAl2022,YuanWengEtAl2021}. Modern deep learning prediction models widely use ego-conditioning, i.e., condition the prediction of adjacent agents' motion on the ego's future motion, to improve the prediction quality and capture the interaction between the ego and the agents. The resulting prediction is then consumed by a planner that aims to generate an ego motion plan that avoids collisions and makes progress towards the goal. Depending on how the prediction is consumed, there are two typical styles of planners: sampling-based planners and iterative planners. The former takes a bunch of ego motion samples, calls the prediction model to generate ego-conditioned predictions and searches for a motion plan \cite{CuiSergioEtAl2021,ChenKarkusEtAl2023}. An iterative planner, on the other hand, iteratively refines the ego motion plan, with e.g. gradient \cite{rosoliaDeBruyneEtAl2016} or Bayesian optimization \cite{CensiCalisiEtAl2008}. While the latter may achieve finer granularity for the ego motion due to iterative refinement, it needs to evaluate the ego motion plan significantly more times than a sampling-based counterpart and the evaluation cannot be parallelized. As a result, the computational complexity prohibits the use of complex deep-learned ego-conditioned prediction models together with an iterative planner---when a prediction model is used, it is typically limited to simple analytical models \cite{ChenRosoliaEtAl2022}. In this paper, we propose a computationally tractable approach, called \fullalgname{} (\algname{}), which reasons about interactivity by combining deep-learned prediction models with iterative planners. IJP significantly outperforms other baselines yielding safer motion plans without sacrificing liveness and being overly conservative.

\textbf{Contributions and paper organization.} We propose \algname{}, which is a model predictive control (MPC)-based planner that is compatible with any (deep learned) prediction model. The two main novelties are (i) \algname{} jointly optimizes over the ego vehicle and the nearby agents' motion with collision avoidance constraints while penalizing deviation from the unconditioned predicted trajectories of the agents. The ``planned" motion for the agents then serve as the ego-conditioned trajectory predictions for those agents and are integrated in the gradient-based planner. (ii) To remedy the local minimum issue of nonconvex optimization, we introduce the novel concept of free-end homotopy that allows us to efficiently explore a diverse range of motions. In particular, free-end homotopy is an extension of homotopy to trajectories that do not share the same end point. We empirically show that \algname{} significantly outperforms a baseline without joint optimization and is superior to a sampling-based planner baseline in both performance and computation complexity.

%% file: related_works.tex
\textbf{Interactive Planning.} Interactive / social-aware planning has been studied extensively in the literature. Some of the early approaches modeled the uncontrolled agents' behavior as Gaussian uncertainty \cite{BatzWatsonEtAl2009} without consideration for the impact of ego behavior on nearby agents. Ignoring the ego's impact can lead to overly conservative motion plans as was famously shown in the freezing robot problem \cite{TrautmanKrause2010}. This led to a plethora of research on navigating crowds while accounting for the reactivity of other agents, such as the joint optimization via Gaussian Process (GP) approach in \cite{TrautmanKrause2010,trautmanmaEtAl2013} and the reinforcement learning (RL) approach in \cite{chenLiuEtAl2019crowd}.

\textbf{Reactive Behavior Modeling.} The crux of interactive planning is to properly model other agents' reactive behavior. Inverse reinforcement learning (IRL) \cite{ZiebartMaasEtAl2008} is an obvious choice, e.g., in \cite{kretzschmarSpiesEtAl2016,kudererGulatiEtAl2015}, and it is used subsequently in optimization over the ego motion \cite{pfeifferSchwesinger2016,SadighSastryEtAl2016c}. Another popular formulation leverages game theory, which assumes that every agent tries to maximize its own utility \cite{bahramLawitzkyEtAl2015,LiOylerEtAl2017}; however, the computational complexity of equilibrium solving remains a challenge and it is not straightforward to combine game theory with data-driven methods. Partially-observable Markov Decision Process (POMDP)-based methods were applied to interactive planning and inferring the hidden intention of surrounding agents \cite{WeiDolanEtAl2013,HubmannBeckerEtAl2017}, but similar to the game-theoretic approaches, POMDPs also suffer from high computational complexity and they are typically hand-crafted, making them difficult to scale. Other analytical models such as Intelligent Driver Models (IDM) \cite{hoermannStumperEtAl2017} and Probabilistic Graphic Models (PGM) \cite{DongDolanEtAl2017} have also been applied to intention estimation and interactive planning, however, they are limited to simple scenarios, such as highway driving. The idea of joint optimization has been studied for conflict resolution \cite{duringPascheka2014}, yet assumes knowledge about the other agent's cooperativeness.  

\textbf{Deep-Learned Prediction.} The above-mentioned methods, though very different in nature, all make assumptions (e.g. rationality) about the surrounding agents' decision processes, and the planner then leverages the assumptions to make the planning problem tractable. In contrast, modern prediction methods are predominantly deep-learned phenomenological models \cite{YuanWengEtAl2021,CasasGulinoEtAl2020b,RhinehartMcAllisterEtAl2019,SalzmannIvanovicEtAl2020,ChenIvanovicEtAl2022}, i.e., models trained with data to match the ground truth without a clear explanation of the decision process. While they achieve good prediction accuracy and are capable of ego-conditioned prediction, working with downstream interactive planner remains difficult, as pointed out previously. The expensive inference of ego-conditioned prediction under a large number of ego plans made it prohibitive to evaluate fine-grained ego plans. In \cite{IvanovicElhafsiEtAl2020} the authors use a linear system to represent the ego-conditioned prediction compactly, but the performance is limited by the simplicity of linear systems.

% Homotopy \cite[Section~1.1]{hatcher2005algebraic} partitions the trajectory space into equivalence classes wherein every member of a class can be continuously ``deformed" to another. 
% \textcolor{red}{[Still working on this paragraph, needs to be shortened and needs a few more papers]} 
% Homotopy \cite[Section~1.1]{hatcher2005algebraic} decomposes the uncountable trajectory space into a countable set of homotopy classes wherein every member of a class can be continuously ``deformed" to another. Thereby, homotopy facilitates efficient search of the trajectory space for motion planning in complex environments with various obstacles. Indeed, 
\textbf{Homotopy Planning.} Homotopy planning has been widely studied for motion planning of autonomous systems. To distinguish among homotopy classes of trajectories, \cite{YiBenderEtAl2018} uses the relative lateral position (i.e., left or right) between two vehicles, \cite{AndersonKarumanchEtAli2012,yi2016homotopy} partition the free space into sub-regions, \cite{bhattacharya2017path,sontges2017computing} construct homotopy-invariant words, while \cite{bhattacharyaLikhachevEtAl2012,bhattacharya2010search} use a magnetic-field inspired approach. All these approaches require the start and end points of all candidate trajectories to coincide for homotopy classes to be well-defined---there exists no concept in the literature on homotopy that accommodates distinguishing trajectories that do not share the same end point. In this paper, we generalize homotopy to rigorously develop the notion of free-end homotopy which provides the same benefits as homotopy to motion planning, but for trajectories that \emph{do not} share the same end point.

% Directly checking Definition~\ref{def:homotopy} to verify if two trajectories belong to the same homotopy class is not straightforward. Multiple approximation methods and work-arounds have been proposed in the literature. \cite{YiBenderEtAl2018} use the relative lateral position (left or right) when two vehicles longitudinal location coincides to determine the homotopy class, yet the criteria is ambiguous as the direction of longitudinal and lateral is not clear in scenarios with curving roads and intersections. In \cite{AndersonKarumanchEtAli2012} the authors partition the free space into non-intersecting polytopes and use the order of region traversing to identify homotopy classes. However, free space partitioning is expensive and only works for static environments. A more common implicit approach is to use multiple trajectory samples as initialization for the gradient-based planner and hoping that one of the solutions is the global optimum \cite{ZuoYangEtAl2020}. However, it is generally inefficient with random initialization as many optimization instances will converge to the same local minimum. 

%% file: homotopy_full.tex
In this section, we will introduce the notion of free-end homotopies that will facilitate faster planning by reducing the number of trajectory initializations for the joint optimization. 

\subsection{Background: Introduction to Homotopy}
% A major caveat of gradient-based optimization for motion planning is its inability to avoid local minima. In particular, unless the optimization problem is convex---which is almost never the case for vehicle motion planning---the solution is not guaranteed to converge to a global minimum. Depending on the initialization, the optimizer may get stuck in a subset of the solution space, e.g., the ego always stays behind a non-ego vehicle, or to its left. These ``isolated" regions of the solution space are sometimes referred to as homotopy classes, which is a concept borrowed from topology. 

Homotopy classes are defined as follows:
% Formally, homotopy classes are equivalence classed defined for a given set of obstacles, and for trajectories that share the same initial and end point.
\begin{definition}[Homotopy Class \cite{bhattacharyaLikhachevEtAl2012}]\label{def:homotopy}
Two continuous trajectories $\mathbf{x}_1:\mathbb{R}\to\mathcal{X}$ and $\mathbf{x}_2:\mathbb{R}\to\mathcal{X}$ connecting the same start and end coordinates $x_s$ and $x_g$, respectively, belong to the same homotopy class \emph{if and only if} one can be continuously deformed into the other without intersecting any obstacles.
\end{definition}
\begin{remark}
    It should be clarified that homotopy is not enforced in an open-ended trajectory optimization problem and it is the limitation of gradient-based optimization that causes the local miminum issue. Nonetheless, studying homotopy offers an intuitive way to partition the solution space into disjoint subsets.
\end{remark}

Directly checking Definition~\ref{def:homotopy} to verify if two trajectories belong to the same homotopy class is not straightforward. Multiple approximation methods and work-arounds have been proposed in the literature. \cite{YiBenderEtAl2018} use the relative lateral position (left or right) when two vehicles longitudinal location coincides to determine the homotopy class, yet the criteria is ambiguous as the direction of longitudinal and lateral is not clear in scenarios with curving roads and intersections. In \cite{AndersonKarumanchEtAli2012} the authors partition the free space into non-intersecting polytopes and use the order of region traversing to identify homotopy classes. However, free space partitioning is expensive and only works for static environments. A more common implicit approach is to use multiple trajectory samples as initialization for the gradient-based planner and hoping that one of the solutions is the global optimum \cite{ZuoYangEtAl2020}. However, it is generally inefficient with random initialization as many optimization instances will converge to the same local minimum.

% The homotopic condition defines an equivalence relationship and we can use it to construct homotopy classes, which are maximum sets of trajectories that within which all members are homotopic to each other.
A key feature of planning for AVs is that the motion plan may not have a fixed end point. For instance, if we require the AV to progress along the road while avoiding obstacles, a particular goal state is not prescribed to the planner. To account for this, we will introduce the notion of free-end homotopy by using magnetic-field homotopy introduced in \cite{bhattacharyaLikhachevEtAl2012}.

\subsection{Introduction to Magnetic-Field-Based Homotopy}
% Generally, two trajectories that belong to the same homotopy class can be continuously deformed from one to the other and therefore it is easier for a gradient-based optimizer to search through one homotopy class. 
% which is well-suited for the key properties exhibited by homotopy planning for autonomous driving:
% \begin{enumerate}
%     % \item The trajectory space of AV is 2-dimensional (X and Y).
%     \item The trajectory space for homotopy planning of AVs is 2-dimensional (X and Y), i.e., the path of the motion plan; speed and orientation along the path are not relevant.
%     \item The planned trajectory has a fixed starting point corresponding to the AV's initial position, but it does not have a fixed ending point.
%     \item AV needs to plan homotopy classes while accounting for multiple moving obstacles in the scenario.
% \end{enumerate}

The magnetic field approach for homotopy class verification \cite{bhattacharyaLikhachevEtAl2012} is based on Ampere's law:
\begin{equation*}
    \oint_C \mathbf{B}\cdot d\mathbf{l} = \mu_0 I_{\text{enc}},
\end{equation*}
which states that the line integral of magnetic field around a closed curve is equal to the product of the magnetic constant $\mu_0$ and the current enclosed $I_{\text{enc}}$. Ampere's law establishes an equivalence condition among all closed curves that encloses the same current, which can also be extended to curves sharing the same starting and ending position. Applying this to homotopy classes in motion planning, the authors in \cite{bhattacharyaLikhachevEtAl2012} let obstacles carry current and calculate the Ampere circuit integral along the robot's trajectory, which is then used to categorize trajectories into different homotopy classes. 

\begin{figure}[t]
    \centering
    \includegraphics[width=0.7\columnwidth]{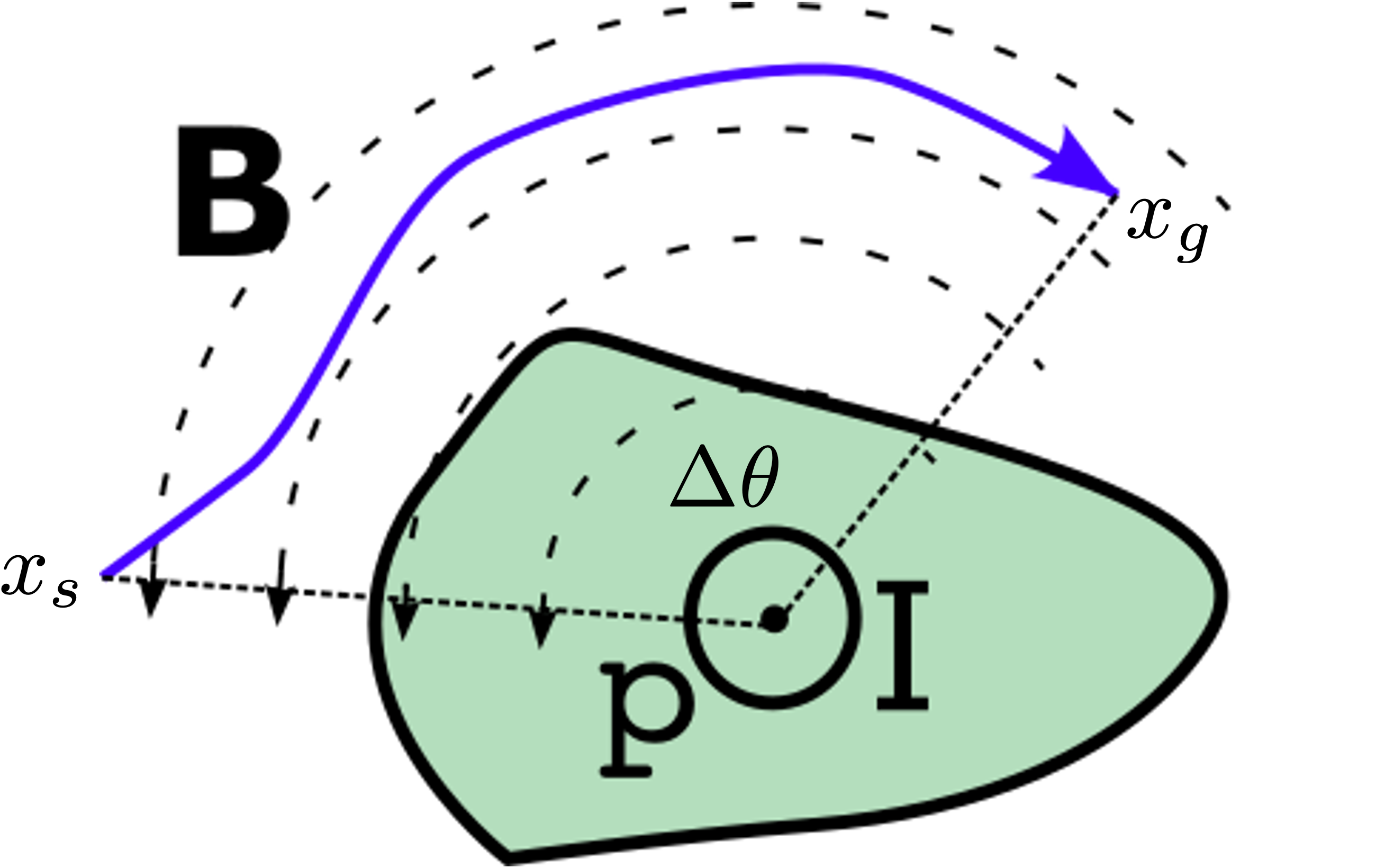}
    \caption{Magnetic path integral in 2D}
    \label{fig:magnetic}
\end{figure}

In 2D space, all obstacles can be viewed as having genus (number of holes) 0 and the imaginary current can be set perpendicular to the X-Y plane crossing the center of the obstacle. Furthermore, the path integral of the magnetic field is easy to compute in 2D; specifically, using Biot-Savart law, the magnetic field near an infinitely long wire at point $p$ with current $I$ perpendicular to the X-Y plane is given by
\begin{equation*}
    \mathbf{B}(\mathbf{r})=\frac{\mu_0 I}{2\pi ||\mathbf{r}||}, 
\end{equation*}
and the direction follows the right-hand law. It follows that the path integral of the magnetic filed along a directional curve that does not intersect with $p$ is simply $\frac{\mu_0I}{2\pi}\Delta \theta$, where $\Delta \theta$ is the angular distance from the start to the end. Fig. \ref{fig:magnetic} illustrates an example where the obstacle is marked in green and the imaginary current that goes through its center $p$, generating a magnetic field $\mathbf{B}$, visualized with the dashed lines. The path integral is then proportional to the angular distance from the start to the end point of the curve. Note that the angular distance is directional and can be negative; when the curve circles $p$ counter-clockwise /clockwise once, the angular distance increases/decreases by $2\pi$, respectively. The angular distance provides two major benefits: (i) it is easy to compute and enforce as a constraint, and (ii) it can be easily extended to moving obstacles, as discussed next.

Let $\mathbf{x}$ be the trajectory of the ego and $\mathbf{x}^{\rm o}$ be the trajectory of an obstacle. To calculate the angular distance $\Delta\theta$, discretize the (X,Y) coordinates of the curves $\mathbf{x}$ and $\mathbf{x}^{\rm o}$ into a sequence of waypoints $\{(X_i,Y_i)\}_{i=1}^N$, $\{(X^{\rm o}_i,Y^{\rm o}_i)\}_{i=1}^N$ and $\Delta \theta$ is computed as
% \begin{equation}\label{eq:angular_dis}
%     \Delta \theta(\mathbf{x},\mathbf{x}^{\rm o}) := \sum_{i=1}^{N-1} \wrap{\arctan \frac{Y_{i+1}-Y^{\rm o}_{i+1}}{X_{i+1}-X^{\rm o}_{i+1}}-\arctan \frac{Y_i-Y^{\rm o}_i}{X_i-X^{\rm o}_i} }
% \end{equation}
\begin{equation}\label{eq:angular_dis}
    \Delta \theta(\mathbf{x},\mathbf{x}^{\rm o}) := \sum_{i=1}^{N-1} \arctan \frac{Y_{i+1}-Y^{\rm o}_{i+1}}{X_{i+1}-X^{\rm o}_{i+1}}-\arctan \frac{Y_i-Y^{\rm o}_i}{X_i-X^{\rm o}_i}
\end{equation}
% where $\wrap{\theta} = ((\theta+\pi) \mod 2\pi)-\pi$ is the angle wrapping function which transforms any $\theta$ to lie within $[-\pi,\pi)$. 
It is clear that \eqref{eq:angular_dis} applies to moving obstacles as well.
% For moving obstacles, $p$ changes over time; however, merely updating $(X_p, Y_p)$ with time in \eqref{eq:angular_dis} is sufficient to deal with the obstacle's motion.

% When the obstacle is moving, $p$ changes with time, one simply discretize $(X_p,Y_p)$ and the angular distance is calculated similarly to \eqref{eq:angular_dis}.

\subsection{Free-end Homotopy} \label{subsec:free-end-homotopy}
As mentioned above, the motion planning problem for AV may not have a fixed end point. Homotopy classes are not well-defined for two curves with different ending points. To resolve this issue, we introduce free-end homotopy, an extension of homotopy, for trajectories that share the same initial point but different end point. The overarching objective is to develop an equivalence class of trajectories, which we call free-end homotopy classes, whose members execute the same relative motion with respect to other agents (e.g., overtake from left of agent 1 and stay behind agent 2) while being continuously transformable to any other member in the class. Free-end homotopy classes facilitate efficient planning by allowing us to downsample motion plan candidates to only those that belong to different free-end homotopy classes, i.e., ones with different relative motions with respect to obstacles.

Let $\mathbf{x}$ be the trajectory of the ego and $\mathbf{x}^{\rm o}$ be the trajectory of a particular obstacle. We begin by defining the \emph{mode} $m:(\mathbf{x},\mathbf{x}^{\rm o})\mapsto m(\mathbf{x},\mathbf{x}^{\rm o})\in\mathbb{Z}$ of a trajectory with respect to a particular obstacle using the angular distances $\Delta\theta$:
% \begin{equation}\label{eq:mode}
%     m(\mathbf{x},\mathbf{x}^{\rm o}):=\left\{ \begin{aligned} 
%     &-1, & \Delta \theta(\mathbf{x},\mathbf{x}^{\rm o}) < -\hat{\theta}\\
%     & 0, & \Delta \theta(\mathbf{x},\mathbf{x}^{\rm o}) \leq |\hat{\theta}|\\
%     & 1, & \Delta \theta(\mathbf{x},\mathbf{x}^{\rm o}) > \hat{\theta}
%     \end{aligned} \right.
% \end{equation}

\begin{figure}
    \centering
    \includegraphics[width=\columnwidth]{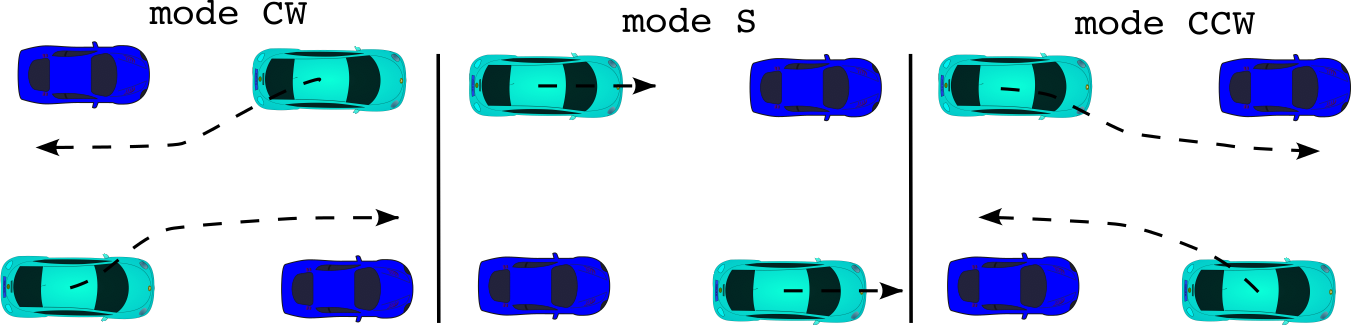}
    \caption{Three homotopy classes: CW, S, and CCW}
    \label{fig:hom_c}
\end{figure}

\vspace{-3mm}
\small
\begin{equation}\label{eq:mode}
    m(\mathbf{x},\mathbf{x}^{\rm o}):=\left\{ \begin{aligned} 
    -(k+1), & & -(\hat{\theta} + k\pi) \leq \Delta \theta(\mathbf{x},\mathbf{x}^{\rm o}) < -(\hat{\theta} + (k+1)\pi)\\
    0, & & -\hat{\theta} \leq \Delta \theta(\mathbf{x},\mathbf{x}^{\rm o}) < \hat{\theta}\\
    k+1, & & \hat{\theta} + k\pi \leq \Delta \theta(\mathbf{x},\mathbf{x}^{\rm o}) < \hat{\theta} + (k+1)\pi
    \end{aligned} \right.
\end{equation}
\normalsize
where $\hat{\theta}$ is a suitably large threshold for differentiating between the three modes.
We refer to these three classes as clockwise (CW), stationary (S), and counter-clockwise (CCW), as illustrated in Fig. \ref{fig:hom_c}. In CW mode, the ego vehicle moves clockwise relative to the object, in CCW the ego vehicle moves counter-clockwise relative to the object, while in S, the ego vehicle remains roughly static relative to the object.

\begin{remark}
Modes with more refined quantization, e.g. consider $\Delta \theta\in [k\pi,(k+1)\pi]$, can be chosen. We chose only three categories as they were found to be sufficient to cover the typical driving scenarios.
\end{remark}

\begin{figure}[t]
    \centering
    \includegraphics{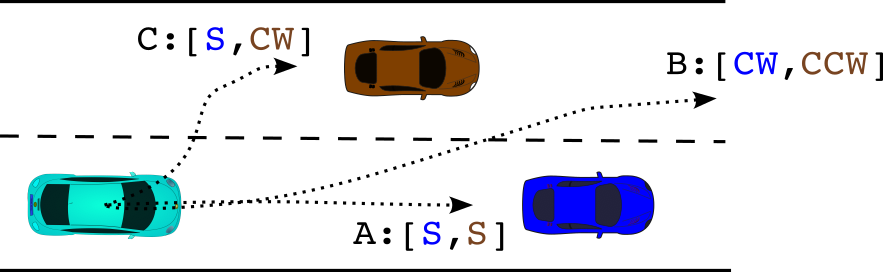}
    \caption{Homotopy classes for two nearby objects where trajectory A is categorized as S for both objects; trajectory B is categorized as CW for the blue car and CCW for the brown car; and trajectory C is categorized as S for the blue car and CW for the brown car.}
    \label{fig:multi_hc}
\end{figure}

If there are $M$ obstacles in the scene, then the \emph{mode vector} $h$ for an ego trajectory $\mathbf{x}$ is defined as the cartesian product of the modes \eqref{eq:mode} with respect to each obstacle in the scene, i.e., $h(\mathbf{x}, \{\mathbf{x}_i^{\rm o}\}_{i=1}^M):=(m(\mathbf{x},\mathbf{x}_1^{\rm o}),\cdots,m(\mathbf{x},\mathbf{x}_M^{\rm o}))$; Fig. \ref{fig:multi_hc} illustrates $h$ with an example scene with two cars near the ego vehicle and three example trajectories. With this, we are now ready to define free-end homotopy. 
\begin{definition}[Free-end Homotopy]\label{def:free-end-homotopy}
    Let $\mathbf{x}_1:\mathbb{R}\to\mathcal{X}$ and $\mathbf{x}_2:\mathbb{R}\to\mathcal{X}$ be two continuous trajectories that share the same start point, but do not necessarily share the same end point. Then, a continuous mapping $f:[0,1]\times\mathbb{R}\to\mathcal{X}$ is called a free-end homotopy if it satisfies the following criteria:
    \begin{itemize}
        \item $f(0,\cdot) = \mathbf{x}_1(\cdot)$, 
        \item $f(1,\cdot) = \mathbf{x}_2(\cdot)$, and
        \item for all $\lambda\in[0,1]$, the mode vector $h_\lambda$ for $f(\lambda,\cdot)$ are equal.
    \end{itemize}
    If a free-end homotopy $f$ exists between $\mathbf{x}_1$ and $\mathbf{x}_2$, then the two trajectories are said to be free-end homotopic.
\end{definition}

Free-end homotopy is a generalization of the notion of homotopy. We make this clear in the next lemma which shows that all homotopic trajectories (with the same start and end points), are also free-end homotopic.
\begin{lemma}\label{lem:homotopy--iff-free-end-homotopy}
    Continuous trajectories with same start and end points are homotopic, \emph{if, and only if}, they are also free-end homotopic. Furthermore, the homotopy is also a free-end homotopy.
\end{lemma}
\begin{proof}
    Proof provided in the appendix.
\end{proof}

% \begin{lemma}\label{lem:homotopy-mode-vector}
%     For continuous trajectories with same start and end points, if the mode vector is the same then they are homotopic.
% \end{lemma}
% \begin{proof}
%     We will prove this by showing the contrapositive: if the trajectories are not homotopic, then they cannot have the same mode vector.
%     Let $\mathbf{x}_1:\mathbb{R}\to\mathcal{X}$ and $\mathbf{x}_2:\mathbb{R}\to\mathcal{X}$ be two continuous trajectories that share the same start and end point, but they are not homotopic. Then, for some obstacle with trajectory $\mathbf{x}^{\rm o}$, we have that $\Delta\theta(\mathbf{x}_1,\mathbf{x}^{\rm o}) \neq \Delta\theta(\mathbf{x}_2,\mathbf{x}^{\rm o})$. However, since the the start and end points are the same, there exists some $k\in\mathbb{Z}\setminus\{0\}$ for which $\Delta\theta(\mathbf{x}_1,\mathbf{x}^{\rm o}) = \Delta\theta(\mathbf{x}_2,\mathbf{x}^{\rm o}) + 2k\pi$ which implies $|\Delta\theta(\mathbf{x}_1,\mathbf{x}^{\rm o}) - \Delta\theta(\mathbf{x}_2,\mathbf{x}^{\rm o})| \geq 2\pi$. Since the modes differ from each other by at most an angle of $\pi$ (see \eqref{eq:mode}), it follows that with a gap of at least $2\pi$, the modes for the two trajectories must be different.
% \end{proof}

In the next lemma we show that the free-end homotopy relation defined above is, in fact, an equivalence relation.
\begin{lemma}[Free-end homotopy is an equivalence relation]\label{lem:eq-relation}
    Free-end homotopy, presented in Definition~\ref{def:homotopy}, is an equivalence relation.
\end{lemma}
\begin{proof}
    Proof provided in the appendix.
\end{proof}
This result ensures that all trajectories that are free-end homotopic can be continuously transformed from one to another while retaining the same mode vector. Hence, we can limit our planning to just one candidate per free-end homotopy class. However, it remains unresolved whether all trajectories with the same mode vector belong to the same free-end homotopy class. Indeed, in the next theorem we will show that only one free-end homotopy class corresponds to one mode vector. This ensures that if we find a continuous trajectory with a particular mode vector, we can \emph{continuously} transform it to any other trajectory with the same mode vector, since they all belong to the same free-homotopy class. This facilitates faster planning by letting us run a continuous optimizer (as discussed later in Section~\ref{sec:method}) on only one trajectory per mode vector.

\begin{thm}[One free-end homotopy class per mode vector]\label{thm:unique-mode-eq-class}
    Continuous trajectories with the same mode vector are free-end homotopic.
\end{thm}
\begin{proof}
    The proof is provided in the appendix. 
    % Here we only provide a sketch. Let $\mathbf{x}_1:\mathbb{R}\to\mathcal{X}$ and $\mathbf{x}_2:\mathbb{R}\to\mathcal{X}$ be two arbitrary continuous trajectories with the same mode vector. We need to construct a free-end homotopy $f$ between these two trajectories. We will do this by constructing a continuous transformation $f_1(\lambda,\cdot)$, that transforms $\mathbf{x}_1$ to $\hat{\mathbf{x}}_1$ (while retaining its mode vector all along the transformation) such that its end point matches the end point of $\mathbf{x}_2$. Then, it follows from \cite{bhattacharyaLikhachevEtAl2012} that two trajectories with the same start and end points and the same mode vector are homotopic. Therefore, there exists a continuous transformation $f_2(\lambda,\cdot)$ that transforms $\hat{\mathbf{x}}_1$ to $\mathbf{x}_2$ while retaining their mode vector. Finally, since $\mathbf{x}_1$ is free-end homotopic to $\hat{\mathbf{x}}_1$ which is free-end homotopic to $\mathbf{x}_2$, by the transitive property of free-end homotopy, $\mathbf{x}_1$ is free-end homotopic to $\mathbf{x}_2$, completing the proof of this theorem.
\end{proof}

\subsection{Applying free-end homotopy classes in motion planning}
% Now that we have a convenient way of identifying free-end homotopy classes for each obstacle in the scene, we need to consider products of homotopy classes. The overall homotopy classes for the ego vehicle is then characterized by the product of the two homotopy classes for the two cars.

When initializing a motion planner, a naive choice is to consider all possible free-end homotopy classes, however, the number of free-end homotopy classes grows exponentially with the number of nearby objects and many of the classes are not realistic. For example, in the situation depicted in Fig. \ref{fig:multi_hc}, CCW for the blue vehicle is not viable as there is not enough space to pass by its right side. For faraway objects, the free-end homotopy class is most likely S due to a small angular distance within the planning horizon.

To identify the promising free-end homotopy classes, we take a sampling approach. Specifically, we use a trajectory sampler to generate $N$ trajectory samples for the ego vehicle. Then we invoke a trajectory predictor to provide scene-centric trajectory predictions for all $M$ objects in the scene. Together there are $N\times M$ free-end homotopy class candidates that are expressed as mode vectors, as described in Section~\ref{subsec:free-end-homotopy}. Among these $N\times M$ mode vectors, many result in repeated mode vectors. Leveraging Theorem~\ref{thm:unique-mode-eq-class}, we only retain the trajectory with the highest reward among all trajectories sharing the same mode vector as a representative for the corresponding free-end homotopy class. The reward function can be any scalar-valued function that scores the performance of the ego trajectory sample amidst the objects' prediction. 

Now that we are left with $K$ trajectories out of the $N\times M$ candidates, each with a unique free-end homotopy class for the whole scene, an ego trajectory sample and predictions for the objects. These $K$ trajectories are used to initialize the gradient-based motion planner in two ways: (i) the nonlinear planning problem is linearized around the ego and the objects' trajectories to create an efficiently-solvable sequential quadratic program (SQP), and (ii) the free-end homotopy class of the trajectory is enforced as a constraint in the planning problem.

%% file: method_full.tex
As discussed in the introduction, modern trajectory planners for autonomous vehicles rely on predictions for nearby agents, and ego-conditioning has been shown to improve the planning performance yet is expensive to run. The proposed \algname{} does not require ego-conditioned predictions, but replaces them with a joint optimization. Specifically, we invoke a prediction module to forecast the non-ego-conditioned future trajectories of the surrounding agents and pass them to the MPC planner.  Intuitively, this prior supplies the optimizer with the non-ego agents' intent. The MPC planner then plans for both the ego vehicle and the surrounding agents to minimize the cost function, which we shall discuss in detail later, while enforcing collision avoidance constraint. In reality, the AV can only control its owm motion, thus assuming control over surrounding agents without any limitation is obviously naive. To remedy this, the cost contains two terms, a term that penalizes nearby agents' deviation from the predicted trajectories, and a term that penalizes their acceleration and jerk. These two terms are interpreted as the price for the ego to force nearby agents away from their nominal path.

\begin{figure}
    \centering
    \includegraphics[width=1\columnwidth]{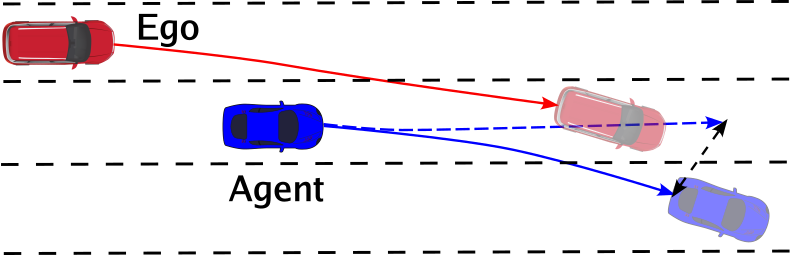}
    \caption{Joint optimization as ego-conditioned prediction: solid trajectories: solution of the joint optimization; dashed line unconditioned predicted trajectory of the blue agent.}
    \label{fig:EC}
\end{figure}

The resulting ``planned" trajectories of nearby agents can be viewed as the ego-conditioned prediction that roughly centers around the unconditioned trajectory prediction. Fig. \ref{fig:EC} illustrates the joint optimization as ego-conditioned prediction. The dashed line shows the unconditioned prediction for the blue agent, which comes from the trajectory predictor; the joint optimization then choose to let the ego (red) change lane and let the blue agent swerve to avoid a collision with the ego, which is viewed as the ego-conditioned prediction, and the deviation from the unconditioned prediction is penalized.
Since the prediction model is only called once without ego-conditioning, the inference time decreases significantly. Moreover, the joint optimization result provides a much finer granularity compared to running ego-conditioned prediction on ego trajectory samples.

Next, we break down the key components of the joint MPC.

\subsection{Dynamic constraints}
We use a Dubin's car model for all vehicles and cyclists in the scene (including the ego vehicle).
\begin{equation}
    x=\begin{bmatrix}X\\Y\\v\\\psi
\end{bmatrix}, u= \begin{bmatrix}\dot{v}\\ \dot{\psi}\end{bmatrix}, x^+=\begin{bmatrix}X+v\cos(\psi)\Delta t\\Y+v\sin(\psi)\Delta t\\v+\dot{v} \Delta t\\\psi+\dot{\psi}\Delta t\end{bmatrix} .
\end{equation}
where $X,Y$ are the longitudinal and lateral coordinates, $v$ and $\dot{v}$ are the longitudinal velocity and acceleration, $\psi$ and $\dot{\psi}$ are the heading angle and yaw rate.

The pedestrians follow a double integrator model with the following dynamics:
\begin{equation*}
    x=\begin{bmatrix}X\\Y\\v_x\\v_y
\end{bmatrix}, u= \begin{bmatrix}\dot{v}_x\\ \dot{v}_y\end{bmatrix}, x^+=\begin{bmatrix}X+v_x\Delta t\\Y+v_y\Delta t\\ v_x+\dot{v}_x \Delta t\\v_y+\dot{v}_y\Delta t\end{bmatrix}.
\end{equation*}

These dynamic models are linearized around an initial guess of $x,u$ pair generated by a trajectory sampler as mentioned in Section \ref{sec:homotopy}. The initial guess satisfies the nonlinear dynamic equations, and the linearized dynamic model takes the form $x^+=Ax+Bu+C$. 

Furthermore, we impose dynamic constraints on state and inputs of the agents. Specifically, for all vehicles,
\begin{align}
    &v\in[v^{\min},v^{\max}]\\
    &|v\dot{\psi}|\le a_y^{\max}\\
    &|\psi|\le \frac{\delta^{\max}}{l} |v| \label{eq:ay_con}\\ 
    &\dot{v}\in[a_x^{\min},a_x^{\max}],
\end{align}
where $[v^{\min},v^{\max}]$ is the velocity range, $a_y^{\max}$ is the maximum lateral acceleration, $a_x^{\min}$ and $a_x^{\max}$ are the lower and upper bounds for longitudinal acceleration, $\delta^{\max}$ is the maximum steering angle and $l$ is the distance between the front and rear axles. All pedestrians follow a simple norm bound on velocity and acceleration:
\begin{equation*}
    ||v||\le v_{\max},\quad ||\dot{v}||\le \dot{v}_{\max}.
\end{equation*}

The dynamic constraints are linearized (especially \eqref{eq:ay_con} so that the effect of velocity is accounted for) and the linearized constraints is written as $G_x^d x+ G_u^d u \le g^d$.

\subsection{Safety constraints}
Safety constraints mainly consist of two parts, collision avoidance constraints and lane boundary constraints. 

\begin{figure}
    \centering
    \includegraphics[width=1\columnwidth]{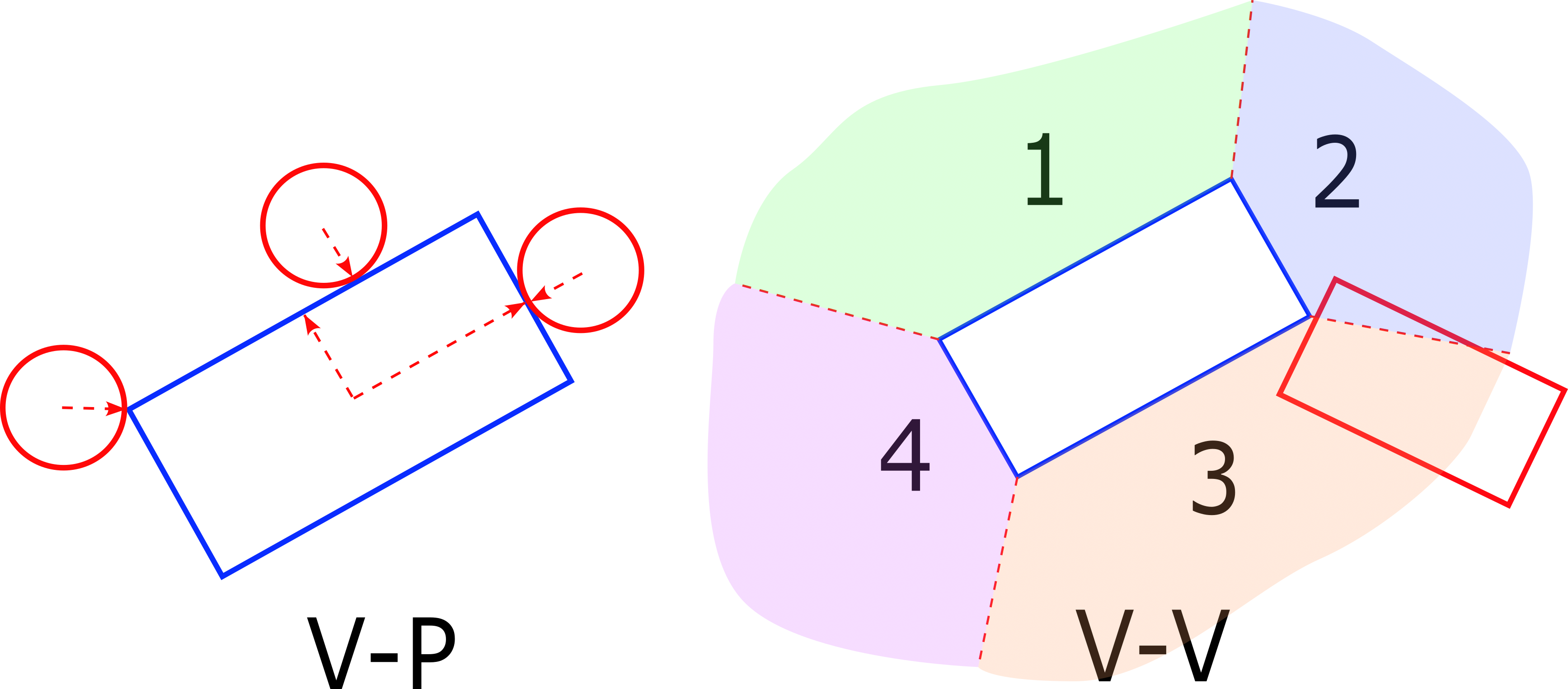}
    \caption{Collision checks between vehicles and pedestrians (left), and two vehicles (right)}
    \label{fig:col}
\end{figure}

All vehicles are modeled as rectangles with varying size (including the ego) and the pedestrians are modeled as a circle with varying radius. The collision avoidance between the ego (rectangle) and pedestrians (circles) is encoded by checking the three cases where the maximum margin is achieved on the X axis, Y axis, and corners of the vehicle, as shown in Fig. \ref{fig:col}. For two vehicles, we analytically calculate the 4 polytopic free spaces around one of the vehicles, as shown in Fig. \ref{fig:col}, and enforce linear constraints that the other vehicle's 4 corners and center point all lie in one of the free spaces. Then we do the same after reversing the role of the two vehicles.

In addition to the collision avoidance constraints, we also enforce the homotopy class constraint, which is computed as discussed in Section \ref{sec:homotopy}. In simulation we observed that the MPC QP behave similarly without the homotopy constraint and the initialization/linearization is sufficient to enforce the homotopy constraint.

\begin{figure}
    \centering
    \includegraphics{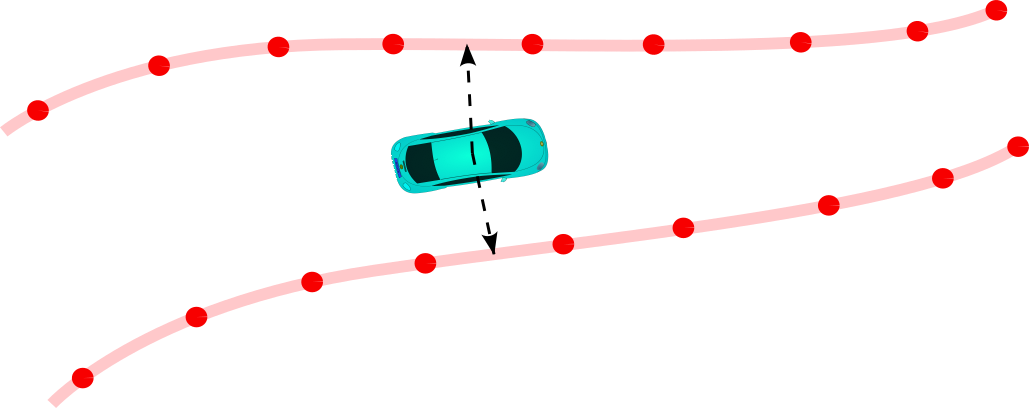}
    \caption{Lane boundary constraint}
    \label{fig:lk_con}
\end{figure}

Lane boundaries are given as polylines (sequence of waypoints with headings), the lane boundary constraints are enforced by projecting the vehicle centers to the polylines and calculate the distance margins, as shown in Fig. \ref{fig:lk_con}.

All of the above mentioned inequality constraints are differentiable w.r.t. the state of the ego vehicle and other agents on the road, and they are linearized and enforced as linear constraints in the MPC. To ensure feasibility, we add slack variable to collision avoidance constraints and lane boundary constraints.

\subsection{Costs and MPC QP setup}
The cost function consists of 5 terms:
\begin{itemize}
    \item Penalty on ego vehicle's tracking error w.r.t. the reference trajectory
    \item Penalty on ego vehicle's acceleration and jerk
    \item Penalty on nearby agents' deviation from their unconditioned trajectory prediction 
    \item Penalty on nearby agents' acceleration and jerk
\end{itemize}

Putting all components together, the joint MPC solves the following QP:
\begin{align}
    \mathop{\min}\limits_{\bue,\buo,\bxe,\bxo} &  \eta_e (\jref(\bxe,\xref)+ \ju(\bue))+\eta_o (\jdev(\bxo,\xpred)+\ju(\buo)) \\
    \mathrm{s.t.}\;& x_e[0]=x_e^0,\;x_o[0]=x_o^0 \label{eq:xinit} \\
    & \forall t=0,...,T-1, i\in\{e,o_1,...o_n\}, \nonumber  \\ 
    & \quad x_i[t+1]=A_i[t]x_i[t]+B_i[t]u_i[t]+C_i[t] \label{eq:dyn_eq} \\ 
    & \quad G_{x,i}^d[t] x_i[t]+G_{u,i}^d[t]u_i[t]\le g^d[t] \label{eq:dyn_ineq}\\ 
    & \forall t=1,...,T, G_e^s[t] x_e[t]+G_o^s[t] x_o[t]\le g^s[t], \label{eq:safety_constr}
\end{align}
where $x_e$ is the future state of the ego vehicle, $x_{o_i}$ is the future state of agent $i$, $A,B,C$ are the matrices corresponding to the dynamic equality constraints, $G_x^d,G_u^d,g^d$ are matrices corresponding to the input and state bounds, $G_e^s,G_o^s,g^s$ define the safety constraints, including collision avoidance, lane boundary, and the homotopy constraint.

The costs include $\jref$ that prompts the ego vehicle to track the desired trajectory, $\ju$ that penalizes acceleration and jerk (both angular and linear), and $\jdev$ that penalizes agents' deviation from their predictions. $\eta_e$ and $\eta_o$ determine the distribution of emphasis on the ego vehicle and the agents. A large $\eta_e$ leads to more selfish and intrusive behavior of the ego and a small $\eta_e$ leads to more altruistic ego behavior.

\subsection{\fullalgname{}}\label{sec:alg}
The \algname{} planner is summarized in Algorithm \ref{alg:MPC}. The inputs are the reference trajectory for the ego vehicle given by some high-level planner, scene context $\mathbf{C}$, lane information $\mathbf{L}$, and the current state of the ego and surrounding agents. 

Firstly, \algname{} calls the trajectory prediction model to generate predictions for the $M$ surrounding agents from the scene context $\mathbf{C}$. \algname{} can work with any prediction model that generates dynamically feasible trajectories for the agents involved. It is preferred that the prediction is scene-centric, i.e., predicting joint trajectories for all agents involved. We use Agentformer \cite{YuanWengEtAl2021} as our default predictor because it is scene-centric, and is shown to work well with the downstream planner in \cite{ChenKarkusEtAl2023}. 

\begin{figure*}
    \centering
    \includegraphics[width=0.85\textwidth]{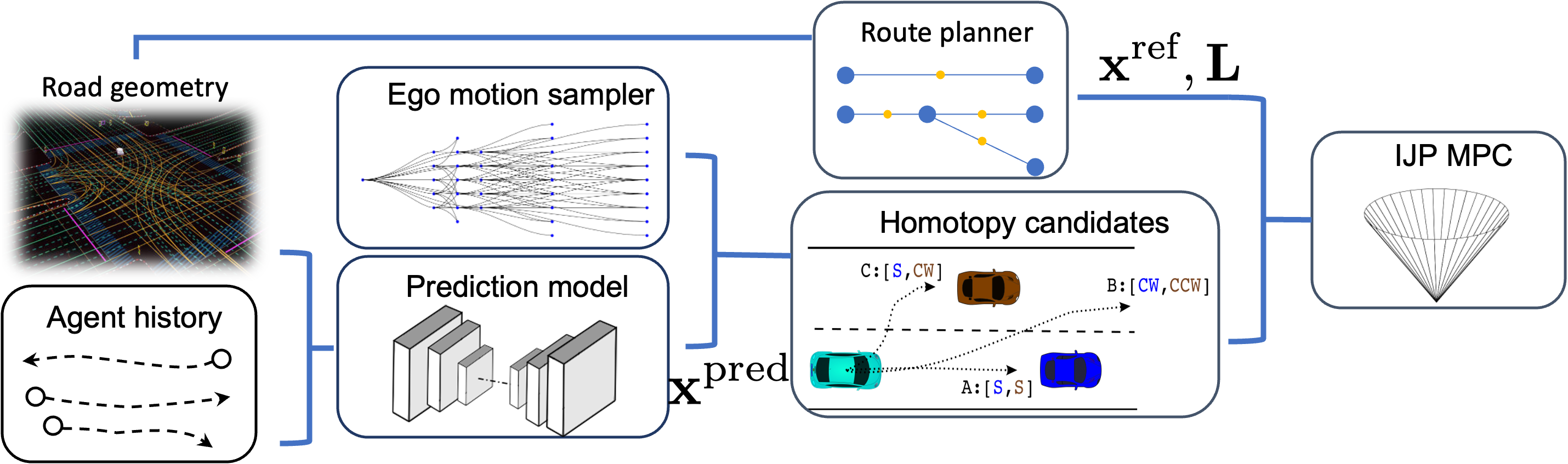}
    \caption{Overview of IJP: the trajectory predictor takes in the lane graph and the agent history to predict the unconditioned prediction for the agents $\xpred$; the route planner plans the desired route and generates the reference trajectory $\xref$ and distills the lane information (such as lane boundaries) $\mathbf{L}$; the trajectory sampler samples the ego trajectory samples, and together with $\xpred$, the homotopy candidates are identified. Finally, \algname{} plans the ego motion via solving the joint model predictive control problem with SQP.}
    \label{fig:overview}
\end{figure*}

\begin{algorithm}
    \caption{\algname{}}
    \label{alg:MPC}
    \begin{algorithmic}[1] % The number tells where the line numbering should start
        \Procedure{\algname{}}{$\xref,\mathbf{C},\mathbf{L},x_e^0,x_o^0$}
            \State $\{\mathbf{x}_{o,j}^{\text{pred}}\}_{j=1}^M\gets \textsc{Traj\_pred}(\mathbf{C})$
            \State $\{\mathbf{x}^{\text{sample}}_{e,i}\}_{i=1}^N\gets\textsc{Ego\_sampling}(x_e^0,\mathbf{L})$ 
            \State $\{(\mathbf{x}_{e,k},\mathbf{x}_{o,k},h_k)\}_{k=1}^K\gets \textsc{Hom\_sel}(\{\mathbf{x}^{\text{sample}}_{e,i}\}_{i=1}^N,\mathbf{x}_o^{\text{pred}})$
            \For{$r=1,...,R$}
                \For {$k=1,...,K$}
                    \State $\text{QP}_k \gets \textsc{Linearize}(\xref,\mathbf{x}_{e,k},\mathbf{x}_{o,k},h_k,x_e^0,x_o^0,\mathbf{L})$
                    \State $\mathbf{x}_{e,k},\mathbf{x}_{o,k}\gets \textsc{Solve\_QP}(\text{QP}_k)$
                \EndFor
            \EndFor
            \State \Return $\bxe$ associated with the best homotopy class.
        \EndProcedure
    \end{algorithmic}
\end{algorithm}

\textsc{Ego\_sampling} takes the ego state and lane information to generate ego trajectory samples with a spline sampler introduced in \cite{ChenKarkusEtAl2023}, which is then used to identify promising homotopies with the predicted trajectories of the surrounding agents in \textsc{Hom\_sel}. With the homotopies selected, \algname{} uses automatic differentiation to linearize the costs, constraints, and dynamics to formulate a quadratic program. JAX \cite{jax2018github} is used for auto-differentiation, and thanks to its powerful parallelization functionality and Just-In-Time (JIT) compilation, the linearization can be done simultaneously for all homotopy classes. The generated QP is solved with 3rd party QP solvers such as GUROBI \cite{gurobi} and Forces Pro \cite{FORCESNLP}. In a sequential quadratic programming (SQP) manner, the nonlinear trajectory optimization problem is linearized and solved as a QP for multiple rounds, each round takes the solution from the last round as the updated linearization point. A proximal constraint is also added to limit the difference of solutions in between rounds to stabilize the SQP.

\begin{remark}
When the trajectory prediction module outputs multimodal predictions of the surrounding agents, the criteria for selecting the optimal solution among the candidate homotopy classes should also take into account the likelihood of the prediction modes, however, we observed that the mode probabilities predicted by the prediction module is usually of bad quality and thus we ignore the mode probability in the final solution selection and simply choose the mode with the lowest cost. We shall investigate how to incorporate prediction likelihood in solution selection in future work.
\end{remark}

%% file: result_full.tex
\subsection{Simulation evaluation setup}
\begin{figure}
    \centering
    \includegraphics[width=0.9\columnwidth]{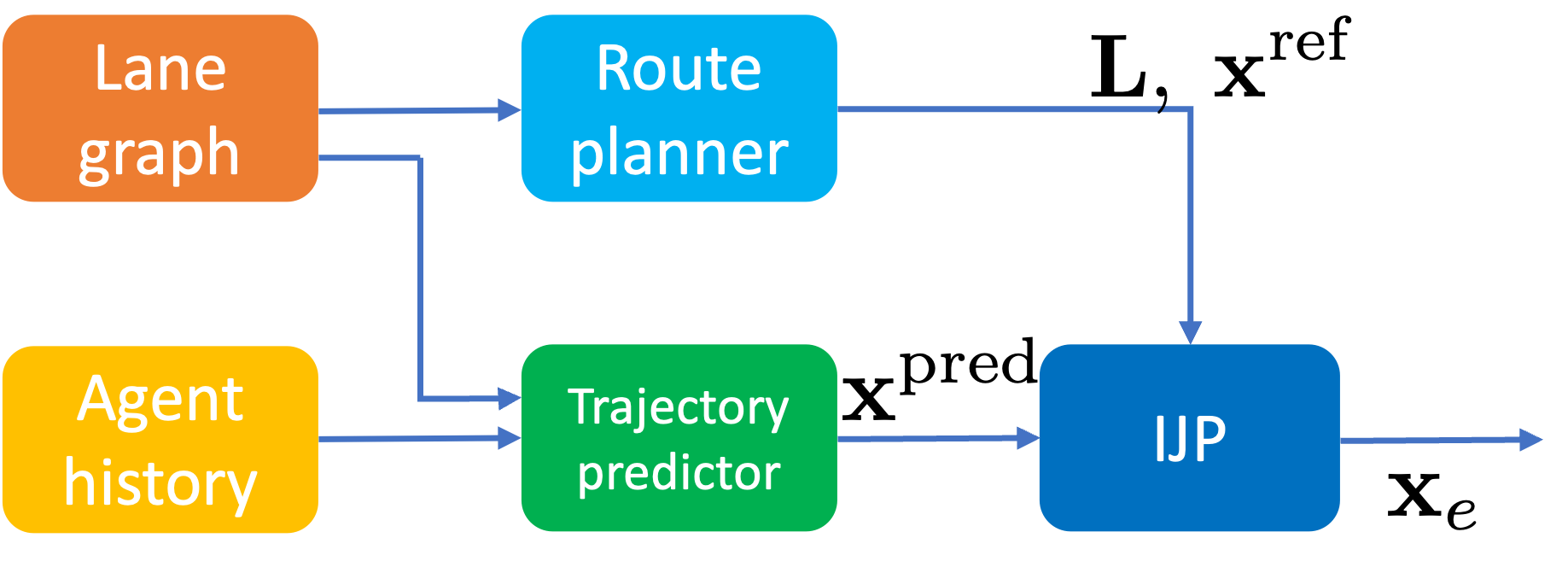}
    \caption{Closed-loop planner structure: the trajectory predictor takes in the lane graph and the agent history to predict the unconditioned prediction for the agents $\mathbf{x}_{\text{pred}}$; the route planner plans the desired route and generates the reference trajectory $\xref$ and distills the lane information (such as lane boundaries) $\mathbf{L}$, and finally \algname{} plans the ego motion.}
    \label{fig:block_diagram}
\end{figure}

We conduct closed-loop simulation in nuPlan \cite{caesarKabzanEtAl2021} to evaluate the proposed approach. The closed-loop planner consists of three modules, a trajectory predictor that generates the unconditioned trajectory prediction, a route planner that distills lane information and reference trajectory from the lane graph, and \algname{} that plans the trajectory, as shown in Fig. \ref{fig:block_diagram}.

We use AgentFormer \cite{YuanWengEtAl2021} as the trajectory predictor without ego-conditioning, which generates 4 samples of predicted future trajectories lasting for 3 seconds.

\noindent\textbf{Route planner. } The route planner takes the lane graph and the ego state as input, and performs a depth-first search to identify the optimal lane sequence. In nuplan simulation, no goal location is provided, instead the lane segments are labeled as "on-route" or "not on-route". The route planner's search criteria is to find the an on-route lane sequence (up to a certain depth) that balances (i) distance to the ego vehicle (ii) length of the lane plan and (iii) total curvature of the lane plan. With a lane sequence selected, the reference trajectory is generated by projecting the ego's current position to the lane centerline and interpolating given the desired ego velocity.

To keep the QP complexity tractable, \algname{} only include a subset of nearby agents in the joint optimization, denoted as EC agents; the rest of the agents are denoted as non-EC agents and \algname{} simply encode collision avoidance constraint with their predicted trajectories. The assignment of EC and non-EC agents is based on their minimum distance to the ego vehicle along their predicted trajectories. When there are less agents than the prescribed number, the MPC QP is padded with dummy agents.

To avoid frequent JIT compilation, the number of EC agents and non-EC agents are fixed so that the MPC QP maintains a fixed problem dimension. 
% One convenient fact is that while the Dubin's car model and double integrator model are different, they share the same state and input dimension. 
When the number of nearby obstacles is larger than the sum of the prescribed number of EC and non-EC agents, far away obstacles are simply ignored.

 We compare the performance of \algname{} to two baselines: (i) non-EC MPC: \algname{} without joint optimization, which only plan the ego behavior and try to avoid collision with the predicted trajectories of nearby agents. (ii) TPP: a sampling-based planner using ego-conditioned prediction, similar to TPP \cite{ChenKarkusEtAl2023} but without multi-layer policy planning.

\begin{remark}
    For fairness of comparison, the non-EC MPC considers a fixed number of non-EC agents, and the number is equal to the sum of EC agents and non-EC agents considered by \algname{}. The TPP planner instead considers all agents detected as the sampling-based algorithm does not require a fixed number of agents.
\end{remark}

\subsection{Simulation result}

\begin{figure}
    \centering
    \includegraphics[width=1\columnwidth]{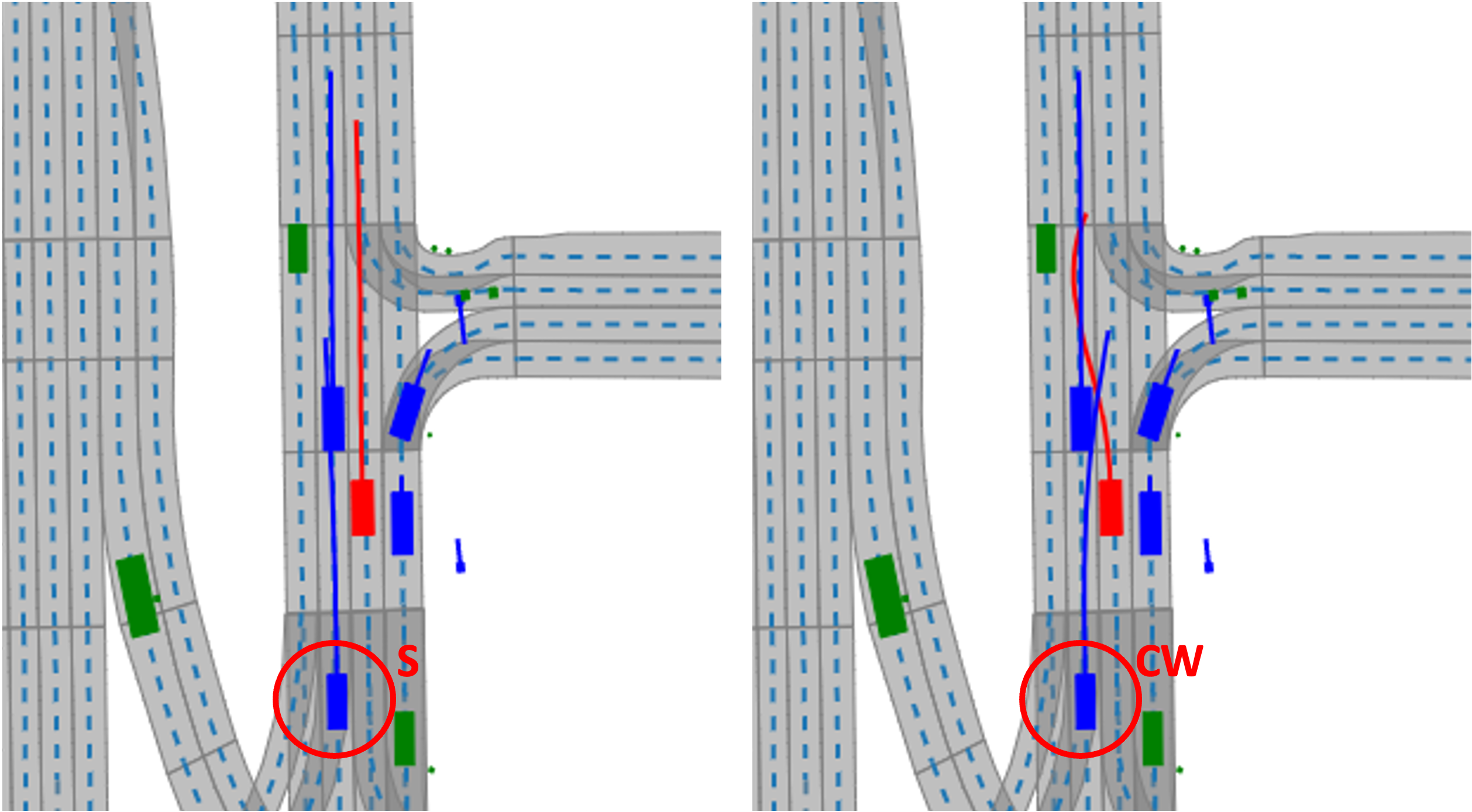}
    \caption{Comparison of the solutions under two homotopy classes: the ego vehicle is in red, the EC agents are in blue and the non-EC agents are in green. As the homotopy class w.r.t. the circled agent changes from S to CW, the ego's behavior changes from lane keeping to lane change, and the "predicted behavior" of the circled vehicle changes accordingly as a result of the joint optimization.}
    \label{fig:EC}
\end{figure}

Fig. \ref{fig:EC} shows an example snapshot from the nuPlan simulation of \algname{} where the two plots are the MPC solutions under two homotopy classes. The only difference between the two homotopy classes is the homotopy w.r.t. the circled vehicle: S (static) in the left case and CW (clockwise) in the right case. The blue curve is the solution of the EC agents' trajectories "planned" by \algname{}. In the right plot, as the ego (red) change lane, the trailing vehicle changes lane to the right to avoid collision with the ego vehicle, which is indeed similar to an ego-conditioned prediction.

Quantitatively, we run closed-loop simulation of 50 scenes from nuPlan's Boston dataset which include many interesting interactive scenarios with sophisticated road geometry. We compare key metrics such as collision rate and progress, all collected from the nuPlan simulator under \algname{} and the baselines, shown in Table \ref{tab:nuplan_metrics}. 

\begin{table}[]

\begin{tabular}{c|cccc}
Planner    & \begin{tabular}[c]{@{}c@{}}Drivable area \\ compliance\end{tabular} & Progress & \begin{tabular}[c]{@{}c@{}}No ego at \\ fault collision\end{tabular} & \begin{tabular}[c]{@{}c@{}}Final\\ Score\end{tabular} \\ \hline
\algname{}        & \textbf{0.98}                                                                & 0.86     & \textbf{0.95}                                                                 & \textbf{0.86}                                                  \\
non-EC MPC & 0.94                                                                & \textbf{0.95}     & 0.66                                                                 & 0.62                                                  \\
TPP        & 0.86                                                                & 0.87     & 0.89                                                                 & 0.71 
\end{tabular}
\caption{Key metrics of in nuPlan simulation}
\label{tab:nuplan_metrics}
\end{table}

The simulation statistics show that \algname{} significantly outperformed the baselines in safety-related metrics such as collision rate and drivable area compliance, and did reasonably well in progress. In fact, upon inspection, we found that the few incidents where the ego was blamed for causing a collision were not correctly assessed. Those few accidents were caused by nearby agents not yielding when making a right turn or lane change. We found no clear mistake made by \algname{} in the 50 scenes in the simulation. The key parameters of \algname{} can be found in Table \ref{tab:param}.

It is counter-intuitive that the non-EC MPC result in worse safety performance given that it fully "respects" the prediction. We suspect that the main reason is when there are multiple agents near the ego vehicle, the prediction makes the motion planning problem infeasible (without slack), and when the prediction is of poor quality, the planner overreact, causing the performance to deteriorate.

Table \ref{tab:compute} shows the computation time of \algname{} and the baselines. Agentformer runs on a Nvidia 3090 GPU and the MPC QP runs on the CPU with Forces Pro QP solver. We separate the build and solve time of the MPC QP because the build process generates all MPC QP instances under different homotopy classes in parallel, while the solve time corresponds to solving one of the QP instances. Comparing to the non-EC MPC, \algname{} takes longer to build and solve as the QP problem is larger. Comparing to TPP, while \algname{} takes longer to solve, it saves more time on the prediction phase as no ego-conditioned prediction is needed.
\begin{remark}
TPP without ego-conditioning would have the same prediction time as \algname{}, but the final score dropped to 0.64 due to more safety violations.
\end{remark} 
\begin{remark}
The computation time of \algname{} can be further improved in at least two ways: parallelizing the solving process of MPC QP under multiple homotopy classes, and utilizing the sparsity pattern in the QP. It is promising that \algname{} can run at a sufficiently high frame rate for real-time planning with these two improvement.
\end{remark}

\begin{table}[]
\begin{tabular}{c|ccc}
           & prediction & Build time & Solve time \\ \hline
\algname{}        & 0.051s       & 0.150s       & 0.152s       \\
non-EC MPC & 0.051s       & 0.045s      & 0.007s     \\
TPP        & 0.690s       & -          & 0.006s     
\end{tabular}
\caption{Computation time of \algname{} and the two baselines}
\label{tab:compute}
\end{table}

\begin{table}[]
\begin{tabular}{ccccc}
\begin{tabular}[c]{@{}c@{}}Number of\\ homotopy\end{tabular} & \begin{tabular}[c]{@{}c@{}}Number of \\ EC agents\end{tabular} & \begin{tabular}[c]{@{}c@{}}Number of \\ non-EC agents\end{tabular} & Horizon & Time step \\ \hline
6                                                            & 6                                                               & 10                                                                  & 3s      & 0.15s    
\end{tabular}
\caption{Key parameters of \algname{}}
\label{tab:param}
\end{table}

%% file: conclusion.tex
We presented a new planning method, \algname{}, that can reason about the impact of the ego's actions on the behavior of other traffic agents by combining gradient-based joint planning for all agents with modern deep learning-based predictors. The key idea behind \algname{} is viewing joint optimization solutions as ego-conditioned predictions and penalizing deviations from the unconditional predictions to regularize the EC predictions. It should be pointed out that the EC predictions currently lack statistical grounding, i.e., no supervision is added in the prediction model training process to force the result of the subsequent joint optimization to match the ego-conditioned ground truth. The main missing piece is counterfactual traffic data, which is not available in general. The behavior of \algname{} largely depends on hyper-parameters such as $\eta_e$ and $\eta_o$, and currently they are hand-tuned. Nonetheless, the closed-loop performance of \algname{} turned out significantly better than the baselines, and we believe the main reasons are the free-end homotopy that diversifies the search space, and the fine granular solution achieved by the joint optimization.

For future work, we will focus on providing a solid probabilistic grounding for the joint optimization solution viewed as ego-conditioned prediction by differentiating through the optimization and training the prediction-planning modules end-to-end.

%% file: appendix.tex
% \subsection{Proofs}
% \begin{remark}\label{rem:time-traj}
%     The time domain of the trajectories can be assumed to be the same without loss of generality (w.l.o.g.). If the times were different, i.e., $T$, $T'$ for $\mathbf{x}_1$ and $\mathbf{x}_2$, respectively, we can scale the coordinates of $\mathbf{x}_2$ to $tT/T'$.
% \end{remark}

\begin{proof}[Proof of Lemma~\ref{lem:homotopy--iff-free-end-homotopy}]
    Let $\mathbf{x}_1:[0,T]\to\mathcal{X}$ and $\mathbf{x}_2:[0,T]\to\mathcal{X}$ be two trajectories such that $\mathbf{x}_1(0)=\mathbf{x}_2(0)$ and $\mathbf{x}_1(T)=\mathbf{x}_2(T)$; the time-domain for both trajectories is chosen to be $T>0$ without loss of generality\footnote{The time domain of the trajectories can be assumed to be the same without loss of generality (w.l.o.g.). If the times were different, i.e., $T$, $T'$ for $\mathbf{x}_1$ and $\mathbf{x}_2$, respectively, we can scale the coordinates of $\mathbf{x}_2$ to $tT/T'$.}. To prove this lemma, we will first show free-end homotopy$\implies$homotopy and then homotopy$\implies$free-end homotopy for $\mathbf{x}_1$ and $\mathbf{x}_2$.
    
    (free-end homotopy$\implies$homotopy) This follows directly by noting that Definition~\ref{def:free-end-homotopy} is stricter than Definition~\ref{def:homotopy} due to the inclusion of the mode vector criteria. Hence, if $\mathbf{x}_1$ and $\mathbf{x}_2$ are free-end homotopic, they satisfy Definition~\ref{def:free-end-homotopy}. Then, $\mathbf{x}_1$ and $\mathbf{x}_2$ also satisfy Definition~\ref{def:homotopy}, implying that they are homotopic.

    (homotopy$\implies$free-end homotopy) Let $\mathbf{x}_1$ and $\mathbf{x}_2$ be homotopic, then there exists a continuous mapping $f:[0,1]\times\mathbb{R}\to\mathcal{X}$ which satisfies the first two criteria of the definition of free-end homotopy in Definition~\ref{def:free-end-homotopy}. We only need to establish the last property in Definition~\ref{def:free-end-homotopy}, i.e., mode vectors for all trajectories along the homotopy transformation by $f$ will be the same.
    To establish this, let $\mathbf{x}^{\rm o}:[0,T]\to\mathcal{X}$ be the trajectory of an arbitrary obstacle in the scene. From \cite[Lemma~3]{bhattacharyaLikhachevEtAl2012} it follows that since the trajectories are homotopic, $\Delta\theta(\mathbf{x}_1,\mathbf{x}^{\rm o}) = \Delta\theta(\mathbf{x}_2,\mathbf{x}^{\rm o})$. Using this in the definition of the mode $m$ \eqref{eq:mode} ensures that $m(\mathbf{x}_1,\mathbf{x}^{\rm o})=m(\mathbf{x}_2,\mathbf{x}^{\rm o})$. Finally, we note that the mode is equal for the two trajectories for an arbitrary obstacle, therefore, it is equal for both trajectories for all obstacles in the scene. Hence the mode \emph{vector} for both the trajectories is the same, completing the proof of this implication, as well as the lemma.
\end{proof}

\begin{proof}[Proof of Lemma~\ref{lem:eq-relation}]
To show that free-end homotopy is an equivalence relation, we must establish that it satisfies the reflexive, symmetric, and transitive properties.
\begin{itemize}
    \item Reflexive: Let $\mathbf{x}:\mathbb{R}\to\mathcal{X}$ have a mode vector $h$. The map $f(\lambda,\cdot) := \mathbf{x}(\cdot)$, which is continuous because $\mathbf{x}(\cdot)$ is continuous, is a free-end homotopy from $\mathbf{x}$ to $\mathbf{x}$.
    \item Symmetric: Let $\mathbf{x}_1:\mathbb{R}\to\mathcal{X}$ and $\mathbf{x}_2:\mathbb{R}\to\mathcal{X}$ be two continuous trajectories that are free-end homotopic. Let $f_{1\to2}(\lambda,\cdot)$ be the free-end homotopy from $\mathbf{x}_1$ to $\mathbf{x}_2$. Then, $f_{2\to1}(\lambda,\cdot):=f_{1\to2}(1-\lambda,\cdot)$ is a free-end homotopy from $\mathbf{x}_2$ to $\mathbf{x}_1$.
    \item Transitive: Let $f_{1\to2}(\lambda,\cdot)$ be the free-end homotopy from $\mathbf{x}_1$ to $\mathbf{x}_2$ and let $f_{2\to3}(\lambda,\cdot)$ be the free-end homotopy from $\mathbf{x}_2$ to $\mathbf{x}_3$. Then, 
    \begin{equation*}
        f_{1\to3}(\lambda, \cdot) := 
        \begin{cases}
            f_{1\to2}(2\lambda, \cdot),~~~\text{if}~~~0\leq \lambda\leq 0.5, \\
            f_{2\to3}(2\lambda-1, \cdot),~~~\text{if}~~~0.5 < \lambda \leq 1,
        \end{cases}
    \end{equation*}
    is a free-end homotopy from $\mathbf{x}_1$ to $\mathbf{x}_3$.
\end{itemize}
This completes the proof of this lemma.
\end{proof}

To prove Theorem~\ref{thm:unique-mode-eq-class}, we first establish the following lemma:
\begin{lemma}\label{lem:homotopy-mode-vector}
    For continuous trajectories with same start and end points, if the mode vector is the same then the trajectories are homotopic.
\end{lemma}
\begin{proof}
    We will prove the contrapositive: if the trajectories are not homotopic, then they cannot have the same mode vector.
    Let $\mathbf{x}_1:\mathbb{R}\to\mathcal{X}$ and $\mathbf{x}_2:\mathbb{R}\to\mathcal{X}$ be two continuous trajectories that share the same start and end point, but they are not homotopic. Then, for some obstacle with trajectory $\mathbf{x}^{\rm o}$, we have that $\Delta\theta(\mathbf{x}_1,\mathbf{x}^{\rm o}) \neq \Delta\theta(\mathbf{x}_2,\mathbf{x}^{\rm o})$. However, since the start and end points are the same, there exists some $k\in\mathbb{Z}\setminus\{0\}$ for which $\Delta\theta(\mathbf{x}_1,\mathbf{x}^{\rm o}) = \Delta\theta(\mathbf{x}_2,\mathbf{x}^{\rm o}) + 2k\pi$ which implies $|\Delta\theta(\mathbf{x}_1,\mathbf{x}^{\rm o}) - \Delta\theta(\mathbf{x}_2,\mathbf{x}^{\rm o})| \geq 2\pi$. Since the modes differ from each other by at most an angle of $\pi$ (see \eqref{eq:mode}), it follows that with a gap of at least $2\pi$, the modes for the two trajectories must be different. This completes the proofs.
\end{proof}

\begin{proof}[Proof of Theorem~\ref{thm:unique-mode-eq-class}]
    Let $\mathbf{x}_1:\mathbb{R}\to\mathcal{X}$ and $\mathbf{x}_2:\mathbb{R}\to\mathcal{X}$ be two arbitrary continuous trajectories with the same mode vector. As we are working with finite-time trajectories, without loss of generality, let the domain of the trajectories be $[0,T]$ where $T>0$. The mode constraint for each obstacle enforces a spatial constraint on the end point of the trajectory. For an arbitrary obstacle, if the mode is $0$, the end point must lie within the convex cone that sweeps an angle of $2\hat{\theta} \leq \pi$ given by the two rays emanating from the obstacle center, while if the mode is a non-zero integer, then the end-point must lie within a convex half-space. To satisfy the mode vector, the end point must, therefore, lie within the intersection $E$ of all these spatial constraint sets; since each of these sets is convex, the intersection set is also convex. As both the trajectories have the same mode vector, their end points $\mathbf{x}_1(T)$ and $\mathbf{x}_2(T)$ lie within $E$. Due to the convexity of $E$, there exists a straight line path $p:[T,2T]\to \mathcal{X}$ defined as $p(t):= (\mathbf{x}_2(T)-\mathbf{x}_1(T))(t-T)/T + \mathbf{x}_1(T)$ for which $p(T)=\mathbf{x}_1(T)$ and $p(2T)=\mathbf{x}_2(T)$ and $p(t)\in E$ for all $t\in[T,2T]$. Now we construct a new trajectory
    \begin{equation*}
        \hat{\mathbf{x}}_1(t) := 
        \begin{cases}
            \mathbf{x}_1(2t),~~~\text{if}~~~0\leq t\leq T/2, \\
            p(2t),~~~\text{if}~~~T/2 < t \leq T.
        \end{cases}
    \end{equation*}
   Let $f_1:[0,1]\times[0,T]\to\mathcal{X}$ be a free-end homotopy candidate from $\mathbf{x}_1$ to $\hat{\mathbf{x}}_1$ as follows:
    \begin{equation*}
        f_1(\lambda,t) := \hat{\mathbf{x}}_1(t(1+\lambda)/2)
    \end{equation*}
    which moves the ending point of $\hat{\mathbf{x}}_1$ from $\hat{\mathbf{x}}_1(T/2)$ to $\hat{\mathbf{x}}_1(T)$. Clearly, $f_1$ is continuous and satsifies the first two criteria in Definition~\ref{def:free-end-homotopy}. Since the end point of $f_1(\lambda,\cdot)$ lies within $E$ for all $\lambda$, it follows that the mode vectors for the trajectories for any $\lambda$ are also the same. Hence, $f_1$ satisfies Defintion~\ref{def:free-end-homotopy}.
    
    We observe that $\hat{\mathbf{x}}_1$ and $\mathbf{x}_2$ have the same mode vector and share the same end points. Using Lemma~\ref{lem:homotopy-mode-vector}, there exists a homotopy $f_2:[0,1]\times[0,T]\to\mathcal{X}$ between them. Furthermore, we know from Lemma~\ref{lem:homotopy--iff-free-end-homotopy} that $f_2$ is also a free-end homotopy. Finally, since $\mathbf{x}_1$ is free-end homotopic to $\hat{\mathbf{x}}_1$, which is free-end homotopic to $\mathbf{x}_2$, by the transitive property of free-end homotopy (Lemma~\ref{lem:eq-relation}), $\mathbf{x}_1$ is free-end homotopic to $\mathbf{x}_2$, completing the proof of this theorem.
\end{proof}